\documentclass[sigconf,preprint]{acmart}

\AtBeginDocument{%
  \providecommand\BibTeX{{%
    \normalfont B\kern-0.5em{\scshape i\kern-0.25em b}\kern-0.8em\TeX}}}

\setcopyright{acmcopyright}
\copyrightyear{2021}
\acmYear{2021}
\acmDOI{}
\acmConference[NUS report]{A technical report by NUS}{April, 2021}{Singapore}
\acmBooktitle{NUS report, April, 2021, Singapore}

\usepackage{graphicx}
\usepackage{subfigure}
\usepackage{algorithm}  
\usepackage{algorithmicx}  
\usepackage{algpseudocode}
\usepackage{multicol}
\usepackage{hyperref}
\usepackage{float}
\begin{document}

\title{An Efficient 2D Method for Training Super-Large Deep Learning Models}

\author{Qifan Xu}
\email{QifanXu@mednet.ucla.edu}
\authornotemark[1]
\affiliation{%
  \institution{University of California, Los Angeles}
  \city{Los Angeles}
  \state{California}
  \country{USA}
}

\author{Shenggui Li}
\email{somerlee.9@gmail.com}
\affiliation{%
  \institution{National University of Singapore}
  \country{Singapore}
  }
  
\author{Chaoyu Gong}
\email{chaoyugong123@gmail.com}
\affiliation{%
  \institution{National University of Singapore}
  \country{Singappore}
}

\author{Yang You}
\email{youy@comp.nus.edu.sg}
\affiliation{%
  \institution{National University of Singapore}
  \country{Singappore}
}

\renewcommand{\shortauthors}{Xu, et al.}

\begin{abstract}

  Huge neural network models have shown unprecedented performance in real-world applications. However, due to memory constraints, model parallelism must be utilized to host large models that would otherwise not fit into the memory of a single device. Previous methods like Megatron partition the parameters of the entire model among multiple devices, while each device has to accommodate the redundant activations in forward and backward pass. In this work, we propose Optimus, a highly efficient and scalable 2D-partition paradigm of model parallelism that would facilitate the training of infinitely large language models. In Optimus, activations are partitioned and distributed among devices, further reducing redundancy. The isoefficiency of Optimus is $W\sim(\sqrt{p}\log p)^3$, which significantly outperforms Megatron’s $W\sim p^3$. On 64 GPUs of TACC Frontera, Optimus achieves 1.48X speedup for training, 1.78X speedup for inference, and 8X increase in maximum batch size over Megatron. Optimus surpasses Megatron in scaling efficiency by a great margin.
  The code is available at \url{https://github.com/xuqifan897/Optimus}.
  
  Mr. Qifan Xu finished this work when he was an intern in Dr. Yang You's group at NUS.

\end{abstract}

\ccsdesc[500]{Computing methodologies~Machine learning}
\ccsdesc[500]{Computing methodologies~Distributed algorithms}
\keywords{matrix multiplication, neural networks, natural language processing}

\maketitle

\section{Introduction}
State-of-the-art deep learning models such as Transformers \cite{transformer} and BERT \cite{BERT} have shown their extraordinary power in the fields of natural language processing and computer vision. By using these well-trained models, realistic images are generated given a short piece of description \cite{DALLE}, and essays can be composed as if they were written by human \cite{megatron}. Behind their impressive performance is the explosion of model size and computational resources \cite{tpu}. The number of parameters in the state-of-the-art NLP models has increased from 110 million (OpenAI GPT) \cite{gpt1} to 340 million (BERT) \cite{BERT}, 1.5 billion (GPT-2) \cite{gpt2}, 8.3 billion (Megatron-LM) \cite{megatron}, 17 billion (Turing-NLP) \cite{deepspeed}, 175 billion (GPT-3) \cite{gpt-3}, 600 billion (Google MoE) \cite{GShard}, and 1.6 trillion (Google Switch Transformer) \cite{switch-tranformer}. Based on this observation, we believe the march of training larger language models will be going on for many years ahead.

Larger models have imposed challenges on both hardware and software. In addition to building supercomputers, novel algorithms are being designed by researchers to tackle these problems. Activation checkpointing \cite{activations} is proposed as a method to trade computation for memory. In checkpointing method, only a portion of intermediate activations are buffered, and the rests are not buffered but recomputed during backward propagation, which can significantly reduce the memory overhead. Mixed precision training \cite{mixed} with dynamic loss scaling replaces 32-bit float tensors with 16-bit half tensors during training, which can reduce the memory overhead and computational cost while preserving the target validation accuracy. 
Efficient optimization strategies for data movement have been shown also highly important for a high-performance transformer implementation \cite{ivanov2020data}.
Moreover, researchers also proposed a method that allows the data to be swapped between GPU and CPU back and forth to fully utilize the computation capability of GPU while extending the memory \cite{zerooffload}. 
These methods are orthogonal to our work.

In addition to these methods, another important paradigm is model parallelism. It is employed to distribute the model parameters among multiple devices. Pipeline parallelism \cite{pipedream, gpipe} is to partition the whole model by layer in a serial manner, so that the input batch is processed on one device at a time, and then sent to the next device. Another parallelism is to partition the parameters of one model to multiple devices. The ZeRO \cite{zero} framework was proposed to split the model horizontally and make each device hold its own partition of parameters and optimizer states, and gather the whole set of parameters of a single model during forward or backward propagation. Leveraging the matrix-matrix multiplication nature of the language models, Megatron \cite{megatron} splits the weight matrices along their rows or columns, and the computation is naturally a block matrix-matrix multiplication. Although parameters are distributed during forward or backward propagation, each device has to accommodate the whole activations and each single layer will return the whole activations after an all-reduce operation. So the activations could become a memory bottleneck when the scale is large. 

It is noted that Megatron essentially adopts a one-dimensional matrix partition strategy in which every single communication involves all the devices. However, with the help of matrix-matrix multiplication algorithms utilizing 2-dimensional partition strategy like SUMMA \cite{van1997summa} and Cannon's algorithm \cite{cannon1969cellular}, it is possible to further speed up both training and inference of huge models.
In this work, we propose Optimus, a novel model parallelism paradigm. Optimus fully distributes parameters and activations, achieving excellent communication efficiency as well as unprecedented memory performance. Optimus is implemented with PyTorch and can be used together with techniques commonly employed in large scale neural network training, as mentioned above.
Optimus is able to achieve an isoefficiency function of $W\sim(\sqrt{p}\log p)^3$, which is much better than that of Megatron ($W\sim p^3$).
We believe the mesh topology of newly emerging supercomputers \cite{tpu} is able to further liberate the unprecedented power of Optimus. The advantages of Optimus over Megatron would help to scale up model training to infinity and facilitate the race of large-scale deep learning models as model size and number of devices continue to increase.
In experiments conducted on TACC Frontera supercomputer, Optimus achieves 1.48$\times$ speedup for training, 1.78$\times$ speedup for a single forward pass, and 8$\times$ increase in maximum batch size over the baseline Megatron model on 64 GPUs. Optimus surpasses Megatron in strong scaling and weak scaling efficiency by a great margin.

In summary, our contributions are as follows:
\begin{itemize}
\item We design and implement an efficient training system for super-large models based on the idea of 2D partition strategy. 
\item We propose several high-performance methods like 2D partition gradient computation, memory-efficient scheme for activations management, and systematic buffering technique.
\item Our methods can be easily implemented by commonly-used frameworks like PyTorch without dealing with compiler-level implementation. The code is available at \url{https://github.com/xuqifan897/Optimus}.
\end{itemize}

\section{Background}
\subsection{Transformer}
From here, we use the following conventions:
\begin{itemize}
\item batch size: $b$
\item sequence length: $s$
\item hidden size: $h$
\item number of attention heads: $n$
\item vocabulary size: $v$
\item number of partitions: $p$
\item SUMMA dimension: $q$
\item number of Transformer layers: $N$
\end{itemize}
We have $p=q^2$ and use a pair of square brackets to express the shape of a tensor.

\begin{figure*}[h]
  \centering
  \includegraphics[width=0.65\linewidth]{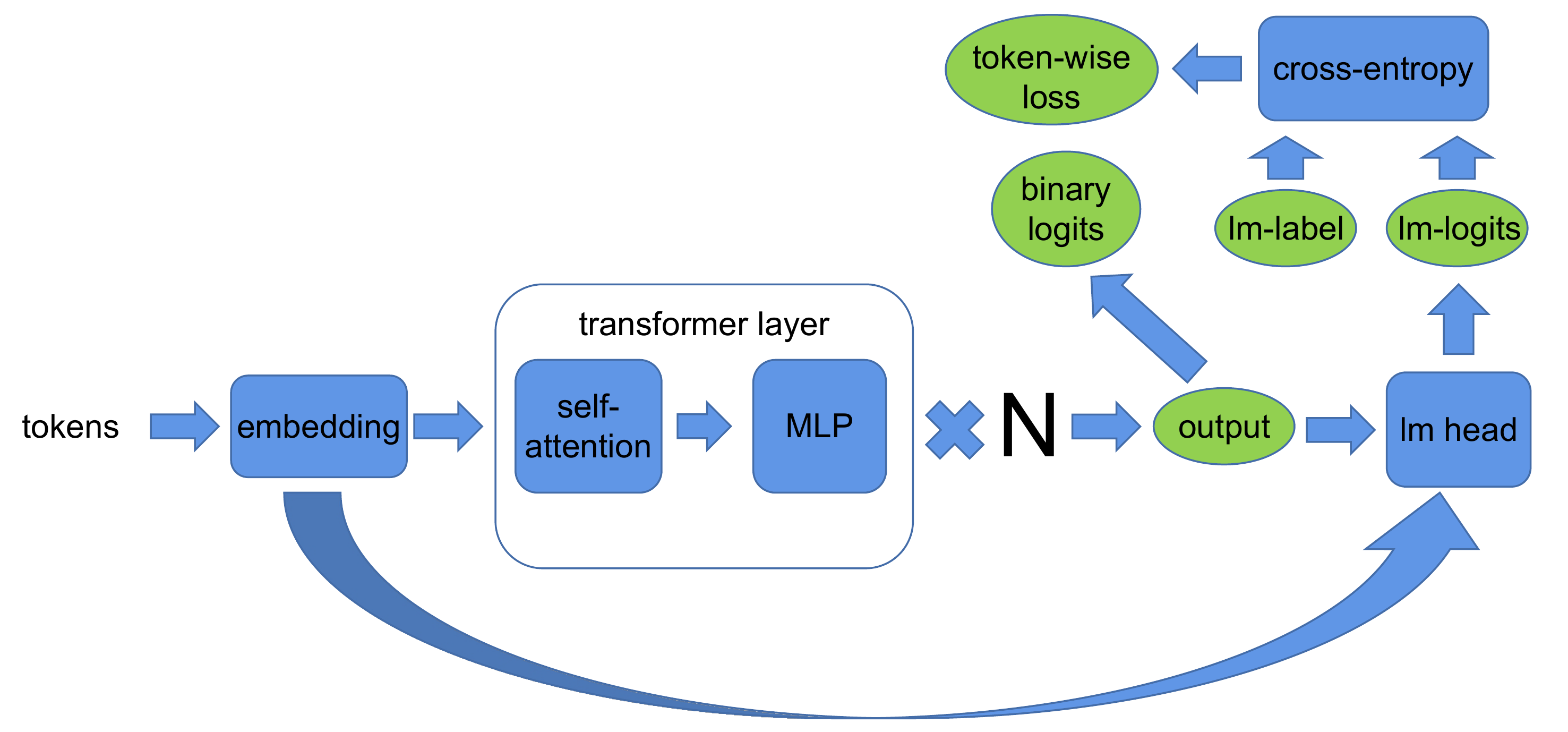}
  \caption{\label{transformer}The structure of Transformer. Blue squares are Transformer modules, green ellipses are inputs, outputs and activations. $N$ is the number of Tranformer layers.}
\end{figure*}

The overall structure of Transformer is shown in Figure \ref{transformer}. Non-linear operations, normalization and residual connections are omitted to highlight the algebraic structure. "Tokens" is an integer tensor of shape $[b,s]$, comprising $b$ sequences of $s$ tokens. Each token is represented by an integer. "Embedding" layer hosts as parameter an embedding table of shape $[v,h]$. "Embedding" layer is to embed every token with a hidden vector. Its output is hidden state, a float tensor of shape $[b,s,h]$. Hidden state is then fed into a sequence of transformer layers. These layers have the same structure but hold distinct parameters. Activations of each layer are of the same shape as inputs, keeping the shape of $[b,s,h]$. And finally, losses are produced in two branches: one branch sends the output to language model head (lm-head), producing the logits at each token position, then calculates the cross-entropy loss with tokenwise labels (lm-labels). This branch is often used in tasks like translation or text generation. The other branch selects the embedding at certain token position, and predicts a binary label for each input sequence. This branch is often used in tasks like sentence classification.

Each Transformer layer consists of a self-attention layer and a multi-layer perceptron. The self-attention layer first multiplies the inputs of shape $[b,s,h]$ with parameters of shape $[h,3h]$, and gets a tensor of shape $[b,s,3h]$, which is then split into query ($Q$), keys ($K$) and values ($V$), all of shape $[b,s,h]$, then subdivided into $n$ attention heads. For a single sequence, $Q$, $K$ and $V$ each can be viewed as $n$ tensors of shape $[s,h/n]$. Attention scores ($A$) $[s,s]$ are calculated as a matrix-matrix multiplication of $Q$ and $K^T$, up to some normalization. $A$ is then multiplied with $V$ to get the output of shape $[s, h/n]$ for a single attention head. The outputs of all attention heads are gathered to get a tensor of shape $[s,h]$, which is then multiplied with parameters of shape $[h,h]$. Finally, the output of the self-attention layer is obtained, of the same shape as inputs.

Multi-layer perceptron (MLP) takes in input activations of shape $[b,s,h]$ and multiplies them with parameters of shape $[h,4h]$, yielding middle activations of shape $[b,s,4h]$, which undergo a nonlinear operation and a linear transformation to recover the shape $[b,s,h]$.

\subsection{Megatron}
\begin{figure*}[h]
	\centering
	\vspace{-0.35cm}
	\subfigure[MLP]{
		\label{megatron_mlp}
		\includegraphics[width=0.8\linewidth]{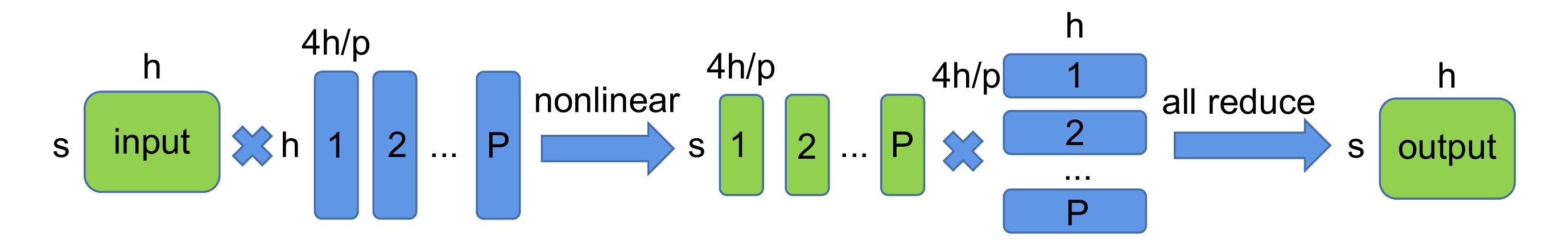}}
	\subfigure[self-attention]{
		\label{megatron_self-attention}
		\includegraphics[width=0.8\linewidth]{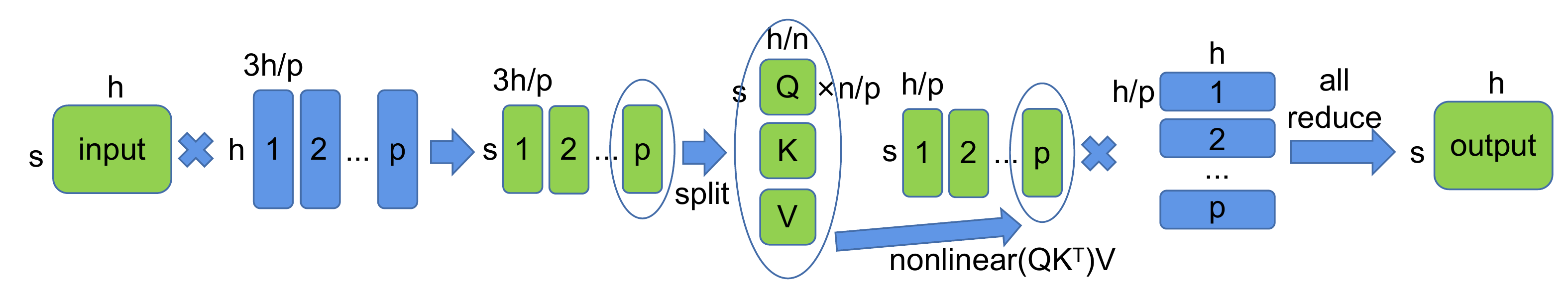}}
	\caption{\label{megatron}Megatron MLP and self-attention. Green squares are 3-dimensional activations with dimension $b$ omitted and blue ones are 2-dimensional parameters. The numbers 1 to $p$ indicate which partition activations or parameters belong to.}
\end{figure*}

The model parallelism employed by Megatron is illustrated in Figure \ref{megatron}, comprising the MLP and self-attention components.

There are two parameter matrices in MLP, the first one is partitioned along its column, and the other one along its row. A copy of inputs is hosted by each device, and is multiplied with one partition of the first matrix on the same device, yielding intermediate activations partitioned along its column. Partitioned intermediate activations on each device are first fed into an activation function and then multiplied with a partition of the second matrix on the same device, yielding outputs of shape $[s,h]$ ($b$ omitted, the same below), the same shape as input. They are all-reduced as the output of MLP.

There are also two parameter matrices in self-attention. The first is partitioned along its column and the other along its row. Inputs are again shared among devices, and are multiplied with one partition of the first parameter matrix on the same device. The activations on each device are partitioned to $Q$, $K$ and $V$. As attention heads are distributed among devices, each device hosts $n/p$ attention heads. So for a single device, each of $Q$, $K$ and $V$ is $n/p$ matrices of shape $[s,h/n]$. $\mathrm{Nonlinear}(QK^T)V$ is calculated within each device, and is reshaped into $[s,h/p]$. They are gathered along $n/p$ attention heads in a single devive and then multiplied with a partition of the second matrix on the same device. All-reduce is performed to yield the output of self-attention.

Communication happens when the all-reduce operation is performed to get the output in the forward pass and to calculate the gradient of input in the backward pass (not shown in Figure \ref{megatron}.) Detailed analysis of communication will be discussed later.

\subsection{Mesh-Tensorflow}
Mesh-Tensorflow \cite{meshTensorflow} is a framework for distributed machine learning, introducing an elegant representation to tensor partition. It first arranges processors into a multi-dimensional mesh. A tensor is either partitioned or replicated along a mesh dimension. Although Mesh-Tensorflow allows multi-dimensional partition, the idea is essentially the same as that of Megatron. Consider a matrix-matrix multiplication: $C=AB$, with $A$ in shape $[\xi, \eta]$, $B$ in shape $[\eta, \zeta]$ and $C$ in shape $[\xi, \zeta]$. Here we prove redundancy is inevitable. In another word, there must be a tensor that is replicated along a mesh dimension. Assume there is no replication of $A$, then both $A$'s two dimensions, $\xi$ and $\eta$, should be partitioned. As a result, $C$ is replicated along the mesh dimension to which the contraction dimension $\eta$ is assigned. Besides, reshape is a communication intensive operation because of the change of the layout of a tensor.

\subsection{SUMMA}
SUMMA stands for Scalable Universal Matrix Multiplication Algorithm. It considers four forms of matrix products, but we only discuss three of them here: $C=AB$, $C=AB^T$ and $C=A^TB$. We notice that these three products form a closed set in terms of differentiation:
\begin{equation}
\label{C=AB}
C=AB\rightarrow A_{grad}=C_{grad}B^T,B_{grad}=A^TC_{grad},
\end{equation}

\begin{equation}
\label{C=ATB}
C=A^TB\rightarrow A_{grad}=BC_{grad}^T,B_{grad}=AC_{grad},
\end{equation}

\begin{equation}
\label{C=AB^T}
C=AB^T\rightarrow A_{grad}=C_{grad}B,B_{grad}=C_{grad}^TA,
\end{equation}
where the superscript $T$ means transpose, and the subsript $grad$ means gradient.

We assume the devices physically form a mesh of dimension $q\times q$. Each matrix is evenly partitioned into $q\times q$ sub-blocks, each assigned to the corresponding device. $A_{ij}$ ($i,j=0,1,...q-1$) denotes the sub-block of $A$ in $i$'s row and $j$'s column. The pseudo-code for $C=AB$, $C=AB^T$ and $C=A^TB$ are shown in algorithms \ref{algoC=AB}, \ref{algoC=ABT} and \ref{algoC=ATB}. It is noted that the mesh layout naturally fits into Google's TPU architecture.

\begin{algorithm}  
\caption{$C=AB$}
\label{algoC=AB}
\begin{algorithmic}[1]
\State \textbf{Input}: $A_{ij}$, $B_{ij}$
\State \textbf{Output}: $C_{ij}$
\State $C_{ij}=0$
\For{$l = 0 \to q-1$}  
	\State broadcast $A_{il}$ within any row
	\State broadcast $B_{lj}$ within any column
	\State $C_{ij}=C_{ij}+A_{il}B_{lj}$
\EndFor  
\State \Return{$C_{ij}$}  
\end{algorithmic}  
\end{algorithm}

\begin{figure*}
\centering
\includegraphics[width=0.8\linewidth]{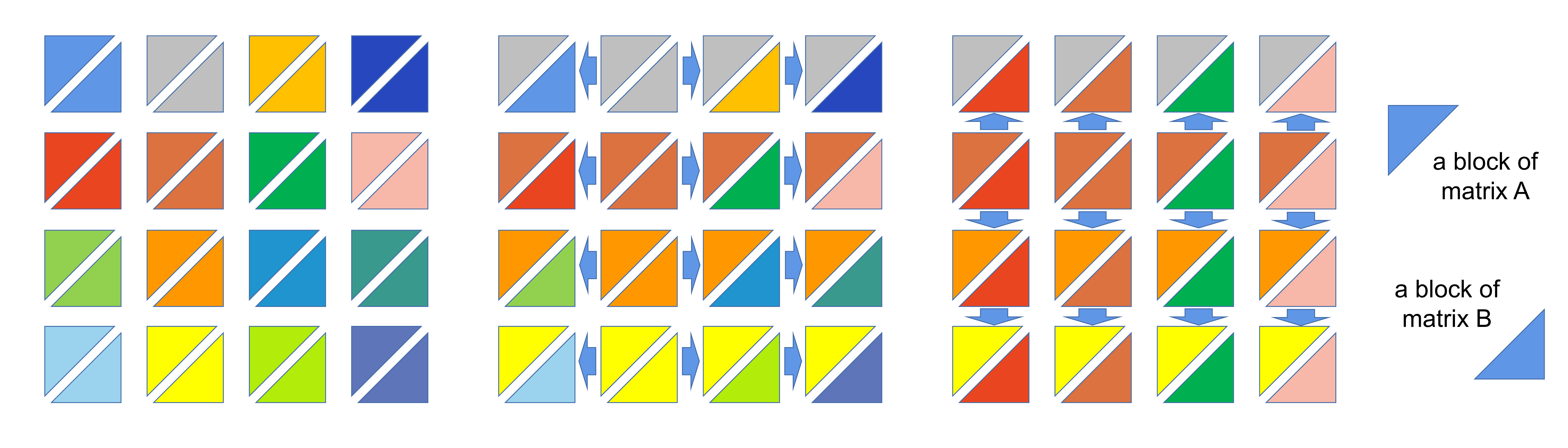}
\caption{\label{SUMMA_diag}An illustration of Algorithm \ref{algoC=AB}. Here we take for example a device mesh of $4\times4$. Different colors represent identities of different devices. At first, each device hosts its own sub-block of matrix $A$ and $B$. Then when computing the outer product $A^2B_2$, each device in the 2nd column broadcast its own sub-block of $A$ along its row, and each device in the 2nd row broadcast its own sub-block of $B$ along its column. Finally, each device performs a local matrix production with the broadcast sub-blocks, and adds it to the final result.}
\end{figure*}

\begin{algorithm}
\caption{$C=AB^T$}
\label{algoC=ABT}
\begin{algorithmic}[1]
\State \textbf{Input}: $A_{ij}$, $B_{ij}$
\State \textbf{Output}: $C_{ij}$
\State $C_{ij}=0$
\For{$l=0 \to q-1$}
	\State broadcast $B_{lj}$ within any column
	\State $C^{temp}_{ij}=A_{ij}(B_{lj})^T$
	\State reduce $C^{temp}_{ij}$ within any row to $C_{il}$
\EndFor
\State \Return{$C_{ij}$}
\end{algorithmic}
\end{algorithm}

\begin{algorithm}  
\caption{$C=A^TB$}
\label{algoC=ATB}
\begin{algorithmic}[1]
\State \textbf{Input}: $A_{ij}$, $B_{ij}$
\State \textbf{Output}: $C_{ij}$
\State $C_{ij}=0$
\For{$l = 0 \to q-1$}  
	\State broadcast $A_{il}$ within any row
	\State $C^{temp}_{ij}=(A_{il})^TB_{ij}$
	\State reduce $C^{temp}_{ij}$ within any column to $C_{lj}$
\EndFor  
\State \Return{$C_{ij}$}  
\end{algorithmic}  
\end{algorithm}

Take Algorithm \ref{algoC=AB} for example, SUMMA perceives matrix-matrix multiplication $C=AB$ as a sum of outer products. If we partition $A$ into columns: $A=[A^1, A^2,...,A^p]$, and $B$ into rows: $B=[B_1, B_2,...B_p]^T$, then $C=\sum_{i=1}^{p}A^iB_i$. We use superscript to denote column-block and subscript to denote row-block here to differentiate them. Next, we only look at the the outer product $A^iB_i$. It turns out that the sub-block of $A^iB_i$ hosted by the device in the $j_{th}$ row and $k_{th}$ column is $A_j^iB_i^k$. And finally one would discover that all devices in the $j_{th}$ row needs $A_j^i$ and all devices in the $k_{th}$ column needs $B_i^k$. A detailed diagram is given in Figure \ref{SUMMA_diag}.

\subsection{Collective communications}
Reduce and broadcast are employed in this work, and are only used within a row or column. The communication costs are determined by the volume of data, bandwidth and latency of the network. We use $\alpha$ to denote the latency, or delay due to data travel time, and $\beta$ to denote the inverse of bandwidth, or time to transfer a scalar. Although the times spent to transfer data within a node and between nodes are different in real-world cases, we do not differentiate them for simplicity here. The time spent in broadcast and reduce in a group of $q$ devices is:
\begin{equation}
T_{\mathrm{broadcast}}=T_{\mathrm{reduce}}=\log(q)\beta B,
\label{T_broadcast}
\end{equation}
where $T_{\mathrm{broadcast}}$ and $T_{\mathrm{reduce}}$ are time spent in broadcast and reduce respectively, and $B$ is the volume of data in a single communication. Latency $\alpha$ is negligible and omitted, as is the same in the following.

Ring all-reduce is employed in Megatron, after which all devices in a group get the same reduced tensor. The time spent in all-reduce within a group of $p$ devices is given below:
\begin{equation}
T_{\mathrm{all-reduce}}=\frac{2\beta(p-1)B}{p}
\label{T_allreduce}
\end{equation}
As shown in the Equations \ref{T_broadcast} and \ref{T_allreduce}, it seems that all-reduce can scale better than reduce and broadcast, even though reduce can be viewed as a sub-task of all-reduce, and broadcast can be achieved by an all-reduce among the source tensor and zeros of the same size. We can make a rough analysis and reach a conclusion that Optimus scales better than Megatron even with the unfair assumption, as shown in Table \ref{cost}.

\section{Optimus}

\subsection{Insights}
\subsubsection{Memory}
As mentioned above, Megatron holds the whole activations in each Transformer layer and this could become a memory bottleneck. Assume there are $N$ Transformer layers in the language model, each device has to accommodate a memory buffer of size $Nbsh/p$ for distributed activation checkpointing. On the other hand, in a single Transformer layer, each device has to host intermediate activations of size $3bsh$ at least (input and output of self-attention, output of MLP). So when $p>N/3$, intermediate activations would become a memory bottleneck compared to the checkpointed activation. For official Transformer with $N=24$, it's hard to run without the out-of-memory problem for $p>8$ while keeping the model size on each device constant. Based on this observation, it's of high necessity to break this memory bottleneck. As SUMMA can fully-distribute the activations without redundancy, it is suitable to address this problem.

\begin{figure*}
	\centering
	\vspace{-0.35cm}
	\subfigure[MLP]{
		\label{optimus_mlp}
		\includegraphics[width=0.8\linewidth]{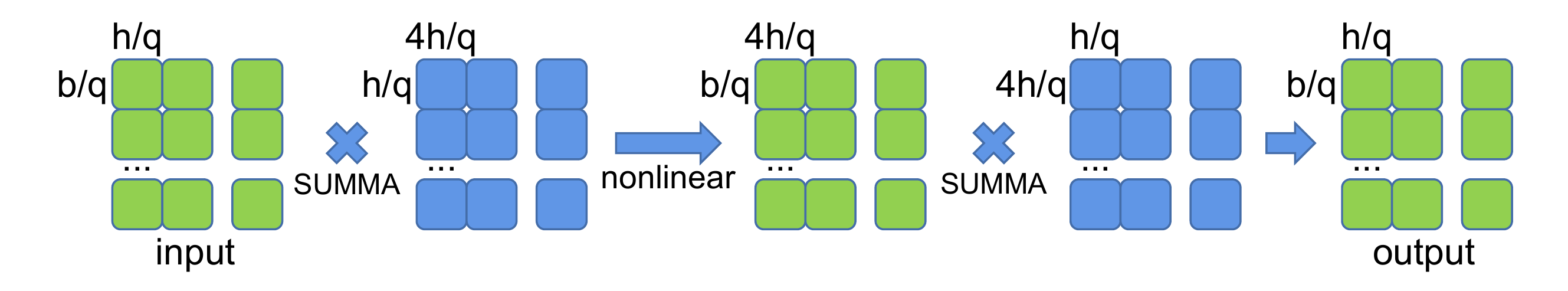}}
	\subfigure[self-attention]{
		\label{optimus_self-attention}
		\includegraphics[width=0.8\linewidth]{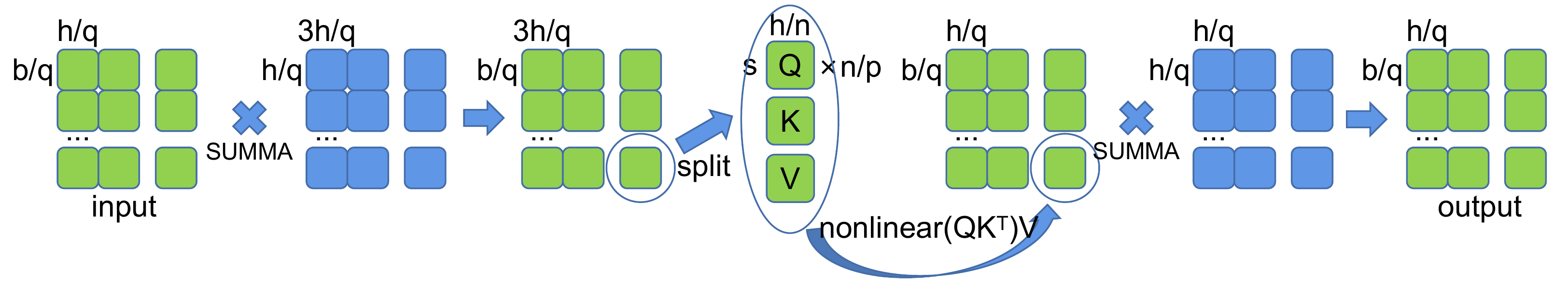}}
	\caption{\label{optimus}The structure of Optimus. Green blocks are activations. All green blocks except $Q$, $K$ and $V$ are three dimensional tensors with $b$ and $h$ split into $q\times q$ sub-blocks, and $s$ omitted. $Q$, $K$ and $V$ are each $n/p$ tensors of shape $[b/q,s,h/n]$, with $b/q$ omitted. Blue blocks are parameter matrices, split in both their two dimensions, into $q \times q$ sub-blocks in SUMMA manner.}
\end{figure*}

\subsubsection{Efficiency}
There are two basic observations in parallel computing:
\begin{enumerate}
\item The overall efficiency decreases as the number of processes increases with a fixed problem size.
\item The efficiency increases as the problem size increases with a fixed number of processes, before running out of memory.
\end{enumerate}
To characterize the scalability of a parallel system, isoefficiency function \cite{isoefficiency} is proposed, which indicates the growth rate of problem size with respect to number of processes to keep the efficiency fixed.

For both Optimus and Megatron, the parallel execution time can be termed as $T_p=W/p+T_{comm}$, where $W$ is the serial execution time, $p$ the number of processes, $T_{comm}$ the communication time. So efficiency can be formulated as $$E=\frac{W}{pT_p}=\frac{1}{1+pT_{comm}/W}.$$

The isoefficiency function is obtained by keeping $pT_{comm}/W$ constant. For the sake of simplicity, we assume $b$ and $n$ scale proportionally to $h$, while $s$ and $N$ are kept constant. So the problem size, or the amount of computation here, is in $O(h^3)$, with MLP dominating the total computation. For Megatron, $T_{comm}$ is the time of all-reduce operations. So we have $$T_{comm}^{Megatron}=2\beta(p-1)\frac{bsh}{p}.$$ And the isoefficiency function for Megatron is $W\sim h^3\sim p^3$.

For Optimus, the communication happens as broadcast and reduce operations. For a mesh of size $q\times q$, the total communication time is in $O(2\beta q\frac{h^2}{p}\log q)$. We can get the isoefficiency function for Optimus: $W\sim h^3\sim(\sqrt{p}\log p)^3$. As Optimus requires a smaller problem size than Megatron to keep the efficiency constant, we can say Optimus scales better than Megatron.

\subsection{Design and Implementation}
\subsubsection{SUMMA-style operations}
The model parallelism of transformer layer employed in Optimus is illustrated in Figure \ref{optimus}. To fit the transformer layer into SUMMA pattern, both the activations and parameters must be uniformly split into $q\times q$ sub-blocks. As a result, activations are also distributed together with parameters in Optimus, while in Megatron, each device has to host whole activations.

MLP in principle included two matrix-matrix multiplications with a non-linear operation in the middle. So it is naturally compatible with SUMMA. Take the first matrix-matrix multiplication for example, it can be viewed as the multiplication of two matrices of shape $[bs,h]$ and $[h,4h]$ respectively, as we do not distinguish dimensions $b$ and $s$ in the multiplication. Activations and the parameter matrix can thus be partitioned into $q\times q$ sub-blocks, each of shape $[bs/q, h/q]$ and $[h/q, 4h/q]$ respectively. The outcome is also distributed to $q\times q$ sub-blocks, each of shape $[bs/q, 4h/q]$. Non-linear operations can be applied locally on each device. The second matrix-matrix multiplication is performed analogously.

As mentioned above, we do not distinguish dimensions $b$ and $s$ in MLP. A natural idea is to partition along the dimensions $s$ and $h$. In this way, $n$ attention heads are each partitioned into $q\times q$ sub-blocks of shape $n\times[b,s/q,h/nq]$. $QK^T$ can be done as described in Algorithm \ref{algoC=ABT} to get the attention scores $A$ distributed into $q \times q$ sub-blocks of shape $n\times [b,s/q,s/q]$. Normalization must be applied within rows as the last dimension is divided. $AV$ can be computed using a similar manner. Although We can get the right result, this method will introduce a huge communication overhead as the total size of $A$ is $bns^2$, which could be much larger than the volume of activations or parameter matrices.

Therefore, we chose to partition along the dimensions $b$ and $h$. In this manner, a whole $s$ is partitioned to one device. This time, each device hosts $Q$, $K$ and $V$ and all of them are in the shape $n/q\times [b/q,s,h/n]$. In other words, each device takes charge of $n/q$ attention heads and $b/q$ sequences. As dimension $s$ is not partitioned, $\mathrm{nonlinear}(QK^T)V$ can be calculated within a single device. A subsequent matrix-matrix multiplication can be again performed in SUMMA manner.

Another important module in Transformer is the embedding layer, which is to assign an embedding vector to every token index. It takes token index matrices of shape $[b,s]$, and outputs embeddings of shape $[b,s,h]$. In principle, this process can be viewed as a matrix-matrix multiplication of a matrix of shape $[b,s,v]$ and a matrix of shape  $[v,h]$. The former matrix is the token index matrix whose last dimension is a one-hot vector indicating the index of the token and the latter matrix is the embedding table. Similarly, it can be easily formulated in SUMMA pattern: the token index matrices are partitioned into $q$ sub-blocks along their rows, each of shape $[b/q,s]$. A row of devices hold identical copies of the the corresponding partition. Moreover, the embedding table is evenly partitioned into $q\times q$ sub-matrices. This partition is compatible with the logits calculation in lm-head, using Algorithm. \ref{algoC=ABT}.

\subsubsection{non-SUMMA-style operations}
\begin{figure}
\centering
\includegraphics[width=\linewidth]{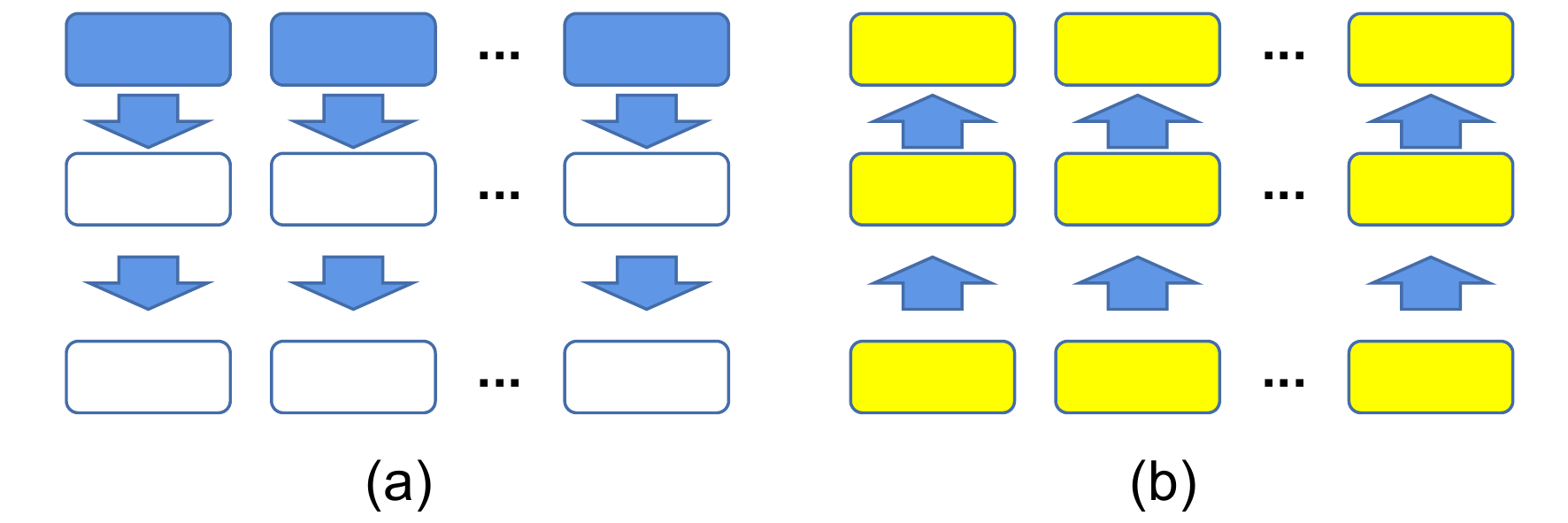}
\caption{\label{bias_add}In operations such as bias-add and layer normalization, parameters are hosted by devices in row 0. (a) In forward pass, Parameters are broadcast along columns. (b) In backward pass, gradients of all devices are reduced to devices in row 0.}
\end{figure}

Besides SUMMA-style matrix-matrix multiplications, Transformer also involves other operations like bias-add. In these operations, parameters are hosted by devices in row 0. In the forward pass, the parameters are broadcast along columns. While in backward propagation, the gradients of the parameters are reduced to the corresponding devices in row 0. This is to ensure that a same parameter is hosted and updated in a single device, and it only introduces limited communication overhead as shown in Figure \ref{bias_add}. Layer normalization is another important operation which needs careful adaptation. Let's put aside the affine transformation, the output of layer normalization is: $$\hat{X}=\frac{X-\mathrm E[X]}{\sqrt{\mathrm{Var}[X]+\epsilon}},$$
where $\mathrm E[X]=\frac1h\sum_{i}X_i$, $\mathrm{Var}[X]=\mathrm{E}[X^2]-(\mathrm{E}[X])^2$ . The gradient of input $X$ is:
$$\frac{1}{\sqrt{\mathrm{Var}[X]+\epsilon}}\left[-\frac1h\left(\sum_j\hat{X}_j\left(\frac{\partial J}{\partial \hat{X}}\right)_j\right)\hat{X} + \frac{\partial J}{\partial \hat{X}} - \frac1h\left(\sum_j\left(\frac{\partial J}{\partial \hat{X}}\right)_j\right)\right].$$
In forward, $X$ and $X^2$ are first summed locally then all-reduced along rows. $\hat{X}$ is calculated with $X$ and $X^2$ and saved together with $1/\sqrt{\mathrm{Var}[X]+\epsilon}$ for backward before being returned as output; while in backward, $\hat{X}\frac{\partial J}{\partial \hat{X}}$ and $\frac{\partial J}{\partial \hat{X}}$ are also treated this way. The communication here are negligible compared to that in SUMMA.

Cross entropy is utilized to calculate the token-wise loss function, and can be implemented analogously. The original form of cross entropy is $H(p,q)=-\sum_{i=1}^{v}p_i\log q_i$, where $p$ and $q$ are ground-truth and predicted distributions respectively. For one-hot ground truth, we assume $p_l=1$, $p_{i,i\neq l}=0$, and the loss reduces to $H(p,q)=-\log q_l$ where $q_l$ is the softmax of input logits $x$. Therefore, we can obtain: 
$$H=\log \left(\sum_{i=1}^ve^{x_i}\right)-x_l$$ 
As $x$ spans a whole SUMMA row of $q$ devices, we first get the local sum and then perform an all-reduce operation along a SUMMA row. We use $\sum_{i=1}^ve^{x_i}$ again to calculate the softmax $$q_j=x_j/\sum_{i=1}^ve^{x_i}$$ and save it for backward. The gradient of input can be easily obtained as $x^{grad}_{j,j\neq l}=q_j$, $x^{grad}_l=q_l-1$.

Compared to SUMMA-style operations, the communication overhead for non-SUMMA-style ones are negligible.

\begin{figure*}
\centering
\includegraphics[width=0.8\linewidth]{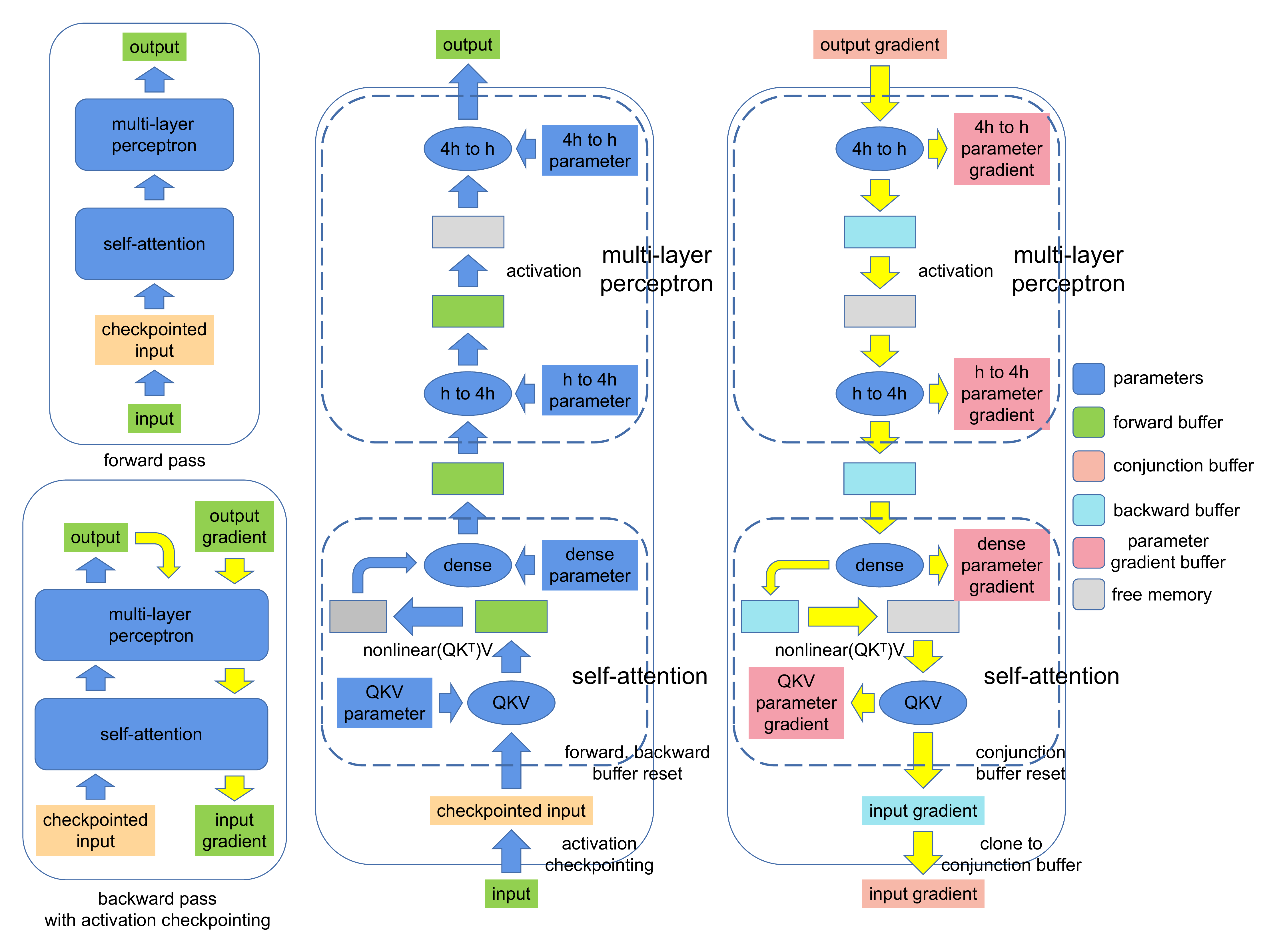}
\caption{\label{checkpointing}For the sack of simplicity, layer normalizations are omitted in the diagram. A forward pass is executed serially (upper left), while in backward pass with activation checkpointing, a forward pass is re-computed before computing the gradients. In forward pass (middle), activations from the previous layer is first checkpointed before forward and backward buffer can be safely reset, as the input is in forward buffer. Outputs of SUMMA-style operations ("QKV", "dense", "h to 4h" and "4h to h") are in the forward buffer, while outputs of activation layer and layer normalization (not shown) are in free memory managed by PyTorch. The output of this Transformer layer is again in forward buffer. In backward pass (right), the output gradient is in conjunction buffer. Gradients of parameters are in parameter gradient buffer, and the input gradients of SUMMA-style operations are in backward buffer. The input gradient of the Transformer layer is first cloned to conjunction layer before passed to the predecessor layer, so that the backward buffer can be safely reset.}
\end{figure*}

\begin{table}
  \caption{Communication and computation costs}
  \label{cost}
  \begin{tabular}{ccc}
    \toprule
    item $\backslash$ scheme&Megatron&Optimus\\
    \midrule
    \shortstack{forward\\communication} & $\frac{4(p-1)}{p}bsh$ & $\frac{\log(p)}{2\sqrt p}(7bsh+12h^2)$\\
		\shortstack{backward\\communication} & $\frac{8(p-1)}{p}bsh$ & $\frac{\log(p)}{2\sqrt p}(21bsh+36h^2)$\\
		\shortstack{forward\\computation} & $\frac{1}{p}(12bsh^2+2bs^2h)$ & $\frac{1}{p}(12bsh^2+2bs^2h)$\\
		\shortstack{backward\\computation} & $\frac{1}{p}(36bsh^2+6bs^2h)$ & $\frac{1}{p}(36bsh^2+6bs^2h)$\\
  \bottomrule
\end{tabular}
\end{table}

\subsubsection{memory pre-allocation and checkpointing}
It is noted that Transformer Layers are identical in structure. To better manage the memory and avoid memory fragmentation, we choose to manually manage the reusable memory In SUMMA-style operations, including workspace buffer, forward buffer, backward buffer, parameter gradient buffer and conjunction buffer.

As indicated in algorithms. \ref{algoC=AB}, \ref{algoC=ABT} and \ref{algoC=ATB}, broadcast and reduce operations require frequent temporary memory allocations to clone the parameters and receive the broadcast or reduced tensors. This would cause memory fragmentation and thus degrade memory performance. Inspired by activation checkpointing, we pre-allocate a piece of memory as a workspace to serve as the temporary memory in the communications. It is noted that matrix-matrix multiplications happen sequentially. So it suffices to allocate the largest volume of memory among those required.

With distributed activation checkpointing, it is sufficient to run a forward and a backward pass for a single Transformer layer. This makes it possible to manually manage all the memory involved, including intermediate activations in the forward pass, gradient for activations, gradient for parameter, and the gradient transmission at the interface between consecutive Transformer layers. The order of different operations are carefully managed so that a same memory buffer is allocated to one tensor at a time to avoid unwanted memory conflict. Detailed illustration is given in Figure \ref{checkpointing} and we have come up with three methods to manage the memory usage:

\begin{enumerate}
\item In theory, we could merge the forward buffer and backward buffer, for that the output of a matrix-matrix multiplication is not needed to calculate the gradients of both inputs. So the memory of output can be reused to host its gradient in the backward pass. However, there are also operations that need their outputs to calculate the gradients of their inputs, like layer normalization. So it's tedious to sort operations to the two categories and determine if the memory of the output can be reused to host its gradient.

\item As parameters are updated independently of each other, we could update the parameters immediately after the backward pass of a Transformer layer, and then reset the parameter gradient buffer. So that the parameter gradient buffer can be reused.

\item For operations like matrix-matrix multiplication and convolution, which are often major operations in neural networks, the output is essentially not needed for backward propagation. As a result, we can in principle skip certain forward pass for these operations in the backward pass with activation checkpointing.
\end{enumerate}

In fact, these three methods apply to all neural networks and we believe they would benefit the memory performance by a great margin.

\section{Analysis and comparison}

\begin{figure*}
\centering
\includegraphics[width=\textwidth]{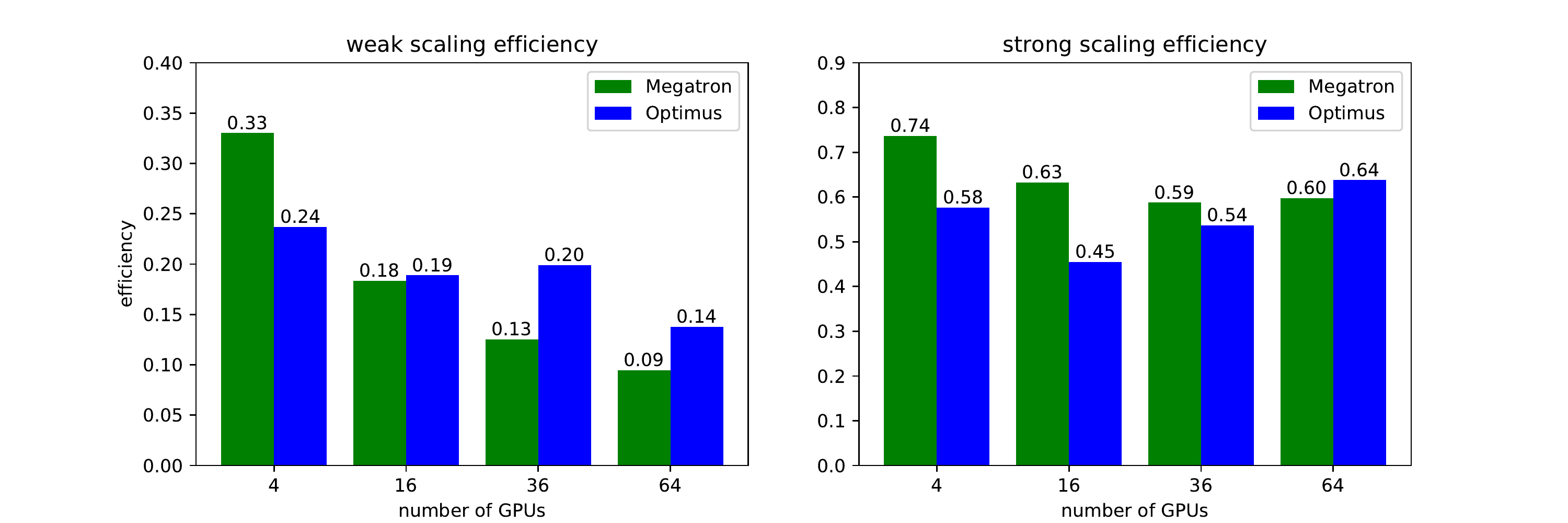}
\caption{Weak scaling and strong scaling.}
\label{scaling}
\end{figure*}

The communication and computation costs are shown in Table \ref{cost}. Expressions in computations are in units of scalar-scalar multiplications, and those of communications are numbers of scalars transferred, ignoring the communication latency. If without activation checkpointing, one backward propagation has twice the computation of a forward pass. With activation checkpointing, one forward pass in backward pass is included. Thus, backward computation is three times forward computation in both Megatron and Optimus. It is noted that backward communication is twice the forward communication in Megatron while it is three times in Optimus. This is because communication happens at the interface of MLP and self-attention layers in Megatron, but it happens together with computation in Optimus. As for computation, it is the same for both Megatron and Optimus as expected.


\begin{figure}[H]
\centering
\includegraphics[width=0.9\linewidth]{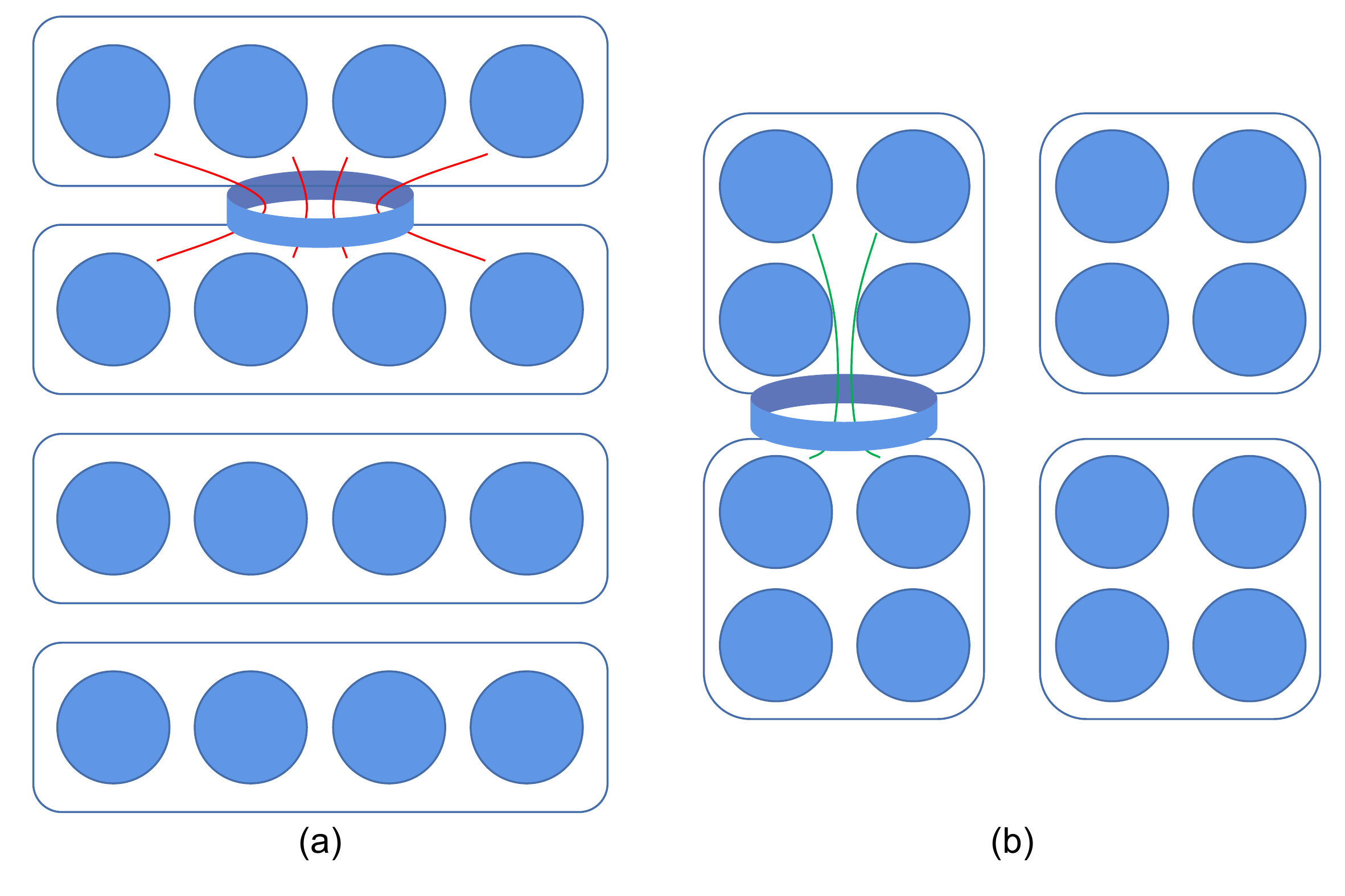}
\caption{\label{rankrearrange}In typical HPC clusters, communication within one node is faster than that spanning multiple nodes. A natural GPU arrangement is shown in (a). Assume we have 4 nodes, each having 4 GPUs. For communication within columns, the data on the 4 GPUs have to crowd on the cable connecting different nodes. Moreover, this communication involves all the 4 nodes. So a better way to arrange the GPUs is shown in (b). In this setting, only two GPUs share the cable, and there are only two nodes involved.}
\end{figure}

\section{Experiments}
Experiments are carried on TACC Frontera rtx nodes. There are 4 NVIDIA Quadro RTX 5000 GPUs on each node, and nodes are interconnected with Mellanox InfiniBand. Experiments are run on 1, 4, 9, 16 nodes respectively. For simplicity, we choose to use the stem of Transformer, or the consecutive Transformer layers, to characterize both communication efficiency and memory performance.
To avoid extensive communication between different nodes, we propose a bunched GPU arrangement as shown in Figure \ref{rankrearrange}.

In this section, we conduct experiments to explore weak scaling \cite{weakscaling} and strong scaling \cite{strongscaling} and use the experiments to demonstrate the superiority of Optimus over Megatron. Strong scaling concerns the speedup for a fixed problem size with respect to the number of processes, and is governed by the Amdahl's law: $\mathrm{speedup}=1/(a+b/p)$, where $a$ is the proportion of execution time spent on the serial part, $b$ is the proportion of execution time spent on the part that can be parallelized, and $p$ is the number of processors. Weak scaling concerns the speedup for a scaled problem size with respect to the number of processors, and is governed by Gustafson’s law: $\mathrm{speedup}=a+b\times p$.

\begin{table*}
	\begin{tabular}{|c|c|c|c|c|c|c|c|c|}
		\multicolumn{9}{c}{Megatron}\\
		\hline
		$\#$nodes & $\#$GPUs & \shortstack{batch\\size} & \shortstack{hidden\\size} & \shortstack{$\#$attention\\heads} & \shortstack{forward time\\/batch size(s)} & \shortstack{backward time\\/batch size(s)} & \shortstack{throughput\\($\#$sequences/s)} & \shortstack{inference\\($\#$sequence/s)}\\
		\hline
		1 & 4 & 60 & 2048 & 32 & 0.0793 & 0.2613 & 2.9363 & 13.1047 \\
		\hline
		4 & 16 & 60 & 4096 & 64 & 0.2081 & 0.5149 & 1.3831 & 4.8046 \\
		\hline
		9 & 36 & 40 & 6120 & 72 & 0.3379 & 0.7955 & 0.8823 & 2.9596 \\
		\hline
		16 & 64 & 30 & 8192 & 128 & 0.4638 & 1.0963 & 0.6410 & 2.1560 \\
		\hline
		\multicolumn{9}{c}{\vspace{0.25cm}}
	\end{tabular}

	\begin{tabular}{|c|c|c|c|c|c|c|c|c|}
		\multicolumn{9}{c}{Optimus}\\
		\hline
		$\#$nodes & $\#$GPUs & \shortstack{batch\\size} & \shortstack{hidden\\size} & \shortstack{$\#$attention\\heads} & \shortstack{forward time\\/batch size(s)} & \shortstack{backward time\\/batch size(s)} & \shortstack{throughput\\($\#$sequences/s)} & \shortstack{inference\\($\#$sequence/s)}\\
		\hline
		1 & 4 & 96 & 2048 & 32 & 0.0985 & 0.2979 & 2.5229 & 10.1502 \\
		\hline
		4 & 16 & 192 & 4096 & 64 & 0.1764 & 0.5312 & 1.4134 & 5.6704 \\
		\hline
		9 & 36 & 288 & 6120 & 72 & 0.1901 & 0.5759 & 1.3055 & 5.2593 \\
		\hline
		16 & 64 & 384 & 8192 & 128 & 0.2589 & 0.7935 & 0.9502 & 3.8625 \\
		\hline
	\end{tabular}
	\caption{\label{comparison}Megatron vs Optimus in weak scaling. Just as in \cite{megatron}, we fix the number of parameters on each GPU. The number of total parameters is proportional to $Nh^2$, and the amount of total computation is $28bsh^2+8bs^2h$, as indicated in Table \ref{cost}. As we fix $N$ to be 24 and $s$ to be $512$, $h$ should be proportional to $q$ or $\sqrt{p}$. Although the amount of total computation is independent of $n$, we still keep $n$ to scale linearly with $p$. Throughput is defined as the ratio between batch size and the sum of forward and backward time, inference is defined as the ratio between batch size and forward time only.}
\end{table*}

\begin{table*}
	\begin{tabular}{|c|c|c|c|c|c|c|c|c|}
		\multicolumn{9}{c}{Megatron}\\
		\hline
		$\#$nodes & $\#$GPUs & \shortstack{batch\\size} & \shortstack{hidden\\size} & \shortstack{$\#$attention\\heads} & \shortstack{forward time\\/batch size(s)} & \shortstack{backward time\\/batch size(s)} & \shortstack{throughput\\($\#$sequences/s)} & \shortstack{inference\\($\#$sequence/s)}\\
		\hline
		1 & 4 & 12 & 3072 & 64 & 0.1225 & 0.4749 & 1.6737 & 8.1616\\
		\hline
		4 & 16 & 12 & 3072 & 64 & 0.1143 & 0.4293 & 1.8397 & 8.7521\\
		\hline
		9 & 36 & 12 & 3096 & 72 & 0.1212 & 0.4512 & 1.7470 & 8.2503\\
		\hline
		16 & 64 & 12 & 3072 & 64 & 0.1195 & 0.5306 & 1.8180 & 8.3711\\
		\hline
		\multicolumn{9}{c}{\vspace{0.25cm}}
	\end{tabular}

	\begin{tabular}{|c|c|c|c|c|c|c|c|c|}
		\multicolumn{9}{c}{Optimus}\\
		\hline
		$\#$nodes & $\#$GPUs & \shortstack{batch\\size} & \shortstack{hidden\\size} & \shortstack{$\#$attention\\heads} & \shortstack{forward time\\/batch size(s)} & \shortstack{backward time\\/batch size(s)} & \shortstack{throughput\\($\#$sequences/s)} & \shortstack{inference\\($\#$sequence/s)}\\
		\hline
		1 & 4 & 24 & 3072 & 24 & 0.1888 & 0.5691 & 1.3195 & 0.4415 \\
		\hline
		4 & 16 & 24 & 3072 & 24 & 0.1950 & 0.5704 & 1.4095 & 5.1285 \\
		\hline
		9 & 36 & 24 & 3072 & 24 & 0.1625 & 0.4764 & 1.5653 & 6.1542 \\
		\hline
		16 & 64 & 24 & 3072 & 24 & 0.1253 & 0.3716 & 2.0123 & 7.9808 \\
		\hline
	\end{tabular}
	\caption{\label{comparison_strong}Megatron vs Optimus in strong scaling. Originally, we want to set $b=24$ and $h=3072$ for both Megatron and Optimus. However, due to memory limit, such batch size could cause out-of-memory for Megatron. So we use $b=12$ for Megatron, as it would not affect the strong scaling efficiency, because both computation and communication are proportional to $b$. Moreover, as Megatron requires $n$ divisible by $p$, while Optimus only requires $n$ divisible by $q$, and they both require $h$ divisible by $n$, we have to set $h$ and $n$ accordingly. As shown in Table \ref{cost}, the change of $n$ does not affect the amount of communication and computation. And increasing $h$ from 3072 to 3096 on 9 nodes for Megatron would have little effect to strong scaling efficiency.}
\end{table*}

\subsection{Weak scaling}
For weak scaling, we fix the number of parameters per device. The specific setting is shown in Tabel \ref{comparison}. We first characterize the time consumption on one GPU, and use it to calculate the theoretical time cost for the other problem sizes, which may not fit in the memory of one GPU. It is noted that due to the fully-distributed nature of Optimus, batch size $b$ can scale linearly with $q$, while we have to decrease the batch size for Megatron to avoid the out-of-memory problem. For Megatron, the time consumption is just proportional to batch size, so the weak scaling efficiency is independent of $b$.

The raw data collected are shown in Table \ref{comparison}. Throughput is the ratio of batch size to the sum of forward and backward time per iteration. Inference is the ratio of batch size to forward time per iteration only. It manifests that Optimus reaches a maximum of 1.4824 times the throughput and 1.7915 times inference compared to Megatron on 64 GPUs. Weak scaling efficiency is shown on the left side of Figure \ref{scaling}. We can observe that the weak scaling efficiency is in a decreasing trend as the number of GPUs increases. That's because the speedup can not grow linearly with the number of processes. However, Optimus can outperform Megatron from 16 GPUs onwards and the advantage of Optimus becomes more prominent as the number of GPUs increases.

\subsection{Strong scaling}
For strong scaling, we fix the problem size, as shown in the Table \ref{comparison_strong}. At first, we want both Megatron and Optimus to have the same settings. However, Megatron requires the number of attention heads to be divisible by the number of GPUs. Thus, we have to make some modification to parameters accordingly while minimizing the change in values. Besides, due to the memory performance of Megatron, it can not host the batch size of 24, so we use the batch size of 12 instead. It is noted that the time consumption is independent of the number of attention heads, and speedup is independent of batch size for Megatron. Again we characterize the time consumption on one GPU. However, due to memory limit, we use half size in each dimension, and calculate the theoretical serial time cost for the settings in Table \ref{comparison_strong}. Strong scaling efficiency is shown on the right side of Figure \ref{scaling}. We can observe that the efficiency of Megatron is in a decreasing trend, while that of Optimus is in an increasing trend, and that Optimus surpasses Megatron on 64 GPUs. The abnormal increasing trend results from the superiority of SUMMA: the communication time decreases as the number of processors increases with a fixed the problem size.

\subsection{memory performance}
\begin{figure}
\centering
\includegraphics[width=\linewidth]{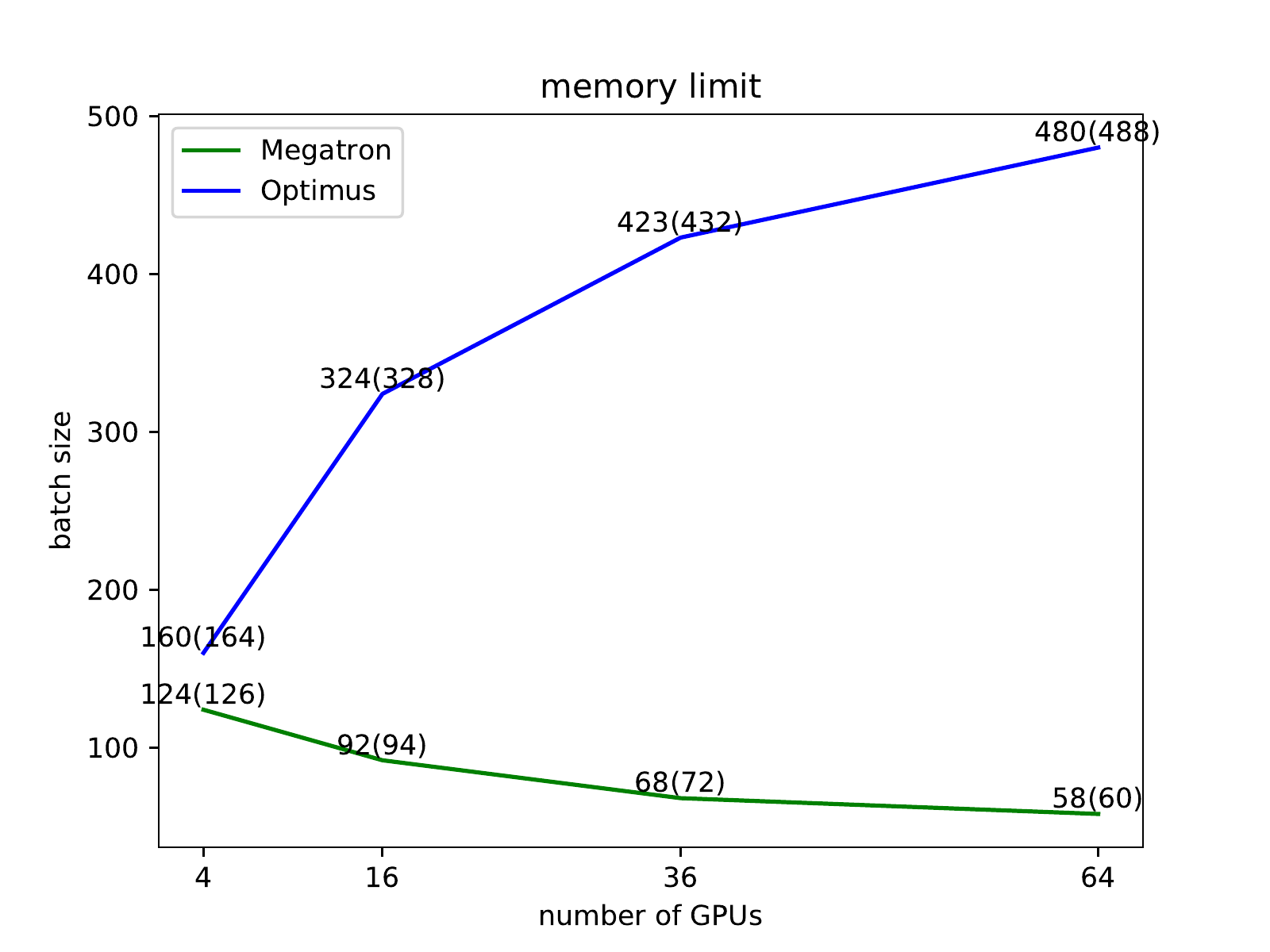}
\caption{\label{memory}The memory limits of Megatron and Optimus. We keep $n$ and $h$ linear with $q$ and fix $N$ and $s$, the same as in Table \ref{comparison}. The data points are labeled in forms of $\xi(\eta)$, which means that Optimus or Megatron can run with $b=\xi$, but can not run with $b=\eta$.}
\end{figure}

For memory performance, Optimus can run at a total batch size of 480 on 64 GPUs. The whole activations would occupy 7.5 GB, which is impossible for Megatron to accommodate. A comparison of memory limits is given in Figure \ref{memory}. It compares the maximum batch size with other parameters the same as in Table \ref{comparison}. We could observe an increasing trend of batch size in Optimus versus a decreasing trend in Megatron. On 64 GPUs, Optimus can run with an amazing batch size 8 times the limit of Megatron.

\section{Conclusion}
In this work, we demonstrate an original model parallelism paradigm which successfully eliminates the memory bottlenecks and increases communication efficiency, thus paving the path for developing infinitely large language models. 
As the power of gigantic models have been demonstrated over and over again, we do not attempt to break the record on metrics like accuracy or perplexity, and leave it for relevant experts.

In addition to the three points given at the end of Section 3, there are other directions for future work. Mixture of Experts (MOE) \cite{MOE} is prevailing and many powerful models are built on top of MOE. We suggest future work to streamline the communication and reduce memory redundancy in such models. Besides, we can manage the memory more thoroughly. In theory, we can manually manage the memory of every operation. Furthermore, operation fusion is another trend. it is noted that in self-attention, the attention scores would in total take up a tensor of shape $[b,n,s,s]$, which would be 8 times the memory of activations of shape $[b,s,h]$, if we take the same scaling in Table \ref{comparison_strong}. However, $\mathrm{Nonlinear}(QK^T)V$ would only take up $bs^2h$ scalar-scalar multiplications, which scales linearly with $p=q^2$ while the computation of MLP scales linearly with $q^3$. So with operation fusion, we can avoid memory allocation for attention scores and other computationally cheap intermediate activations and their gradients, thus further improve the memory performance.



\bibliographystyle{ACM-Reference-Format}
\bibliography{citations}

\end{document}